\documentclass[11pt,a4paper]{article} % For LaTeX2e

\usepackage{arr_latex/acl}
\usepackage{times}

\usepackage[normalem]{ulem}
\usepackage{microtype}
\usepackage{hyperref}
\usepackage{url}
\usepackage{booktabs}
\usepackage{graphicx}
\usepackage{xspace}
\usepackage{enumitem}
\usepackage{amsmath}
\usepackage{amsfonts}
\usepackage{wrapfig}
\usepackage{multicol}
\usepackage{multirow}
\usepackage{lipsum}
\usepackage{algorithm}% http://ctan.org/pkg/algorithm
\usepackage{algpseudocode}
\usepackage{algorithmicx}
\usepackage{inconsolata}

\newcommand{\discern}{\textsc{DiScErN}\xspace}
\newcommand{\discernf}{\discern-F\xspace}
\newcommand{\secref}[1]{\S\ref{#1}}

\title{\discern: Decoding Systematic Errors in Natural Language for Text Classifiers}

\author{Rakesh R. Menon \quad Shashank Srivastava\\
UNC Chapel Hill\\
\texttt{\{rrmenon, ssrivastava\}@cs.unc.edu}
}

\begin{document}

\maketitle

\begin{abstract}
Despite their high predictive accuracies, current machine learning systems often exhibit systematic biases stemming from annotation artifacts or insufficient support for certain classes in the dataset. 
Recent work proposes automatic methods for identifying and explaining systematic biases using keywords. 
We introduce \discern, a framework for interpreting systematic biases in text classifiers using language explanations.
\discern iteratively generates \textit{precise} natural language descriptions of systematic errors by employing an interactive loop between two large language models.
Finally, we use the descriptions to improve classifiers by augmenting classifier training sets with synthetically generated instances or annotated examples via active learning.
On three text-classification datasets, we demonstrate that language explanations from our framework induce consistent performance improvements that go beyond what is achievable with exemplars of systematic bias.
Finally, in human evaluations, we show that users can interpret systematic biases more effectively (by over 25\% relative) and efficiently when described through language explanations as opposed to cluster exemplars.\footnote{Code is available at: \url{https://github.com/rrmenon10/DISCERN}}

\end{abstract}

\section{Introduction}
\label{sec:introduction}
 A broader adoption and trust in machine learning systems would require a confluence of high predictive performance and human interpretability. Despite their high predictive accuracies, current machine learning systems often exhibit systematic biases \citep{robertson2024google, kayser2020google, ulin2018microsoft} stemming from annotation artifacts \citep{gururangan-etal-2018-annotation, mccoy-etal-2019-right} or insufficient support for certain classes in the dataset \citep{Sagawa*2020Distributionally}.
Such biases impede the deployment of systems for real-world applications.
Hence, identifying data \textit{sub-populations} where systems underperform is crucial for a comprehensive understanding of its 
limitations, thereby guiding future refinement strategies.

In line with this objective, to identify \textit{semantically meaningful} sub-populations 
 whose examples have similar characteristics because of a shared underlying structure, previous work 
 proposes to cluster examples and qualitatively examine clusters where the system performs poorly \citep{d2022spotlight}.
In efforts to alleviate the necessity for manual analysis, recent works propose automatic methods to identify and explain underperforming clusters 
by associating keywords with underperforming clusters 
\citep{eyuboglu2022domino, jain2023distilling, deim2023hua}.
However, identifying relevant keywords requires domain expertise and even then, they may not capture all error types.
\begin{figure*}
    \centering
    \includegraphics[width=0.99\textwidth]{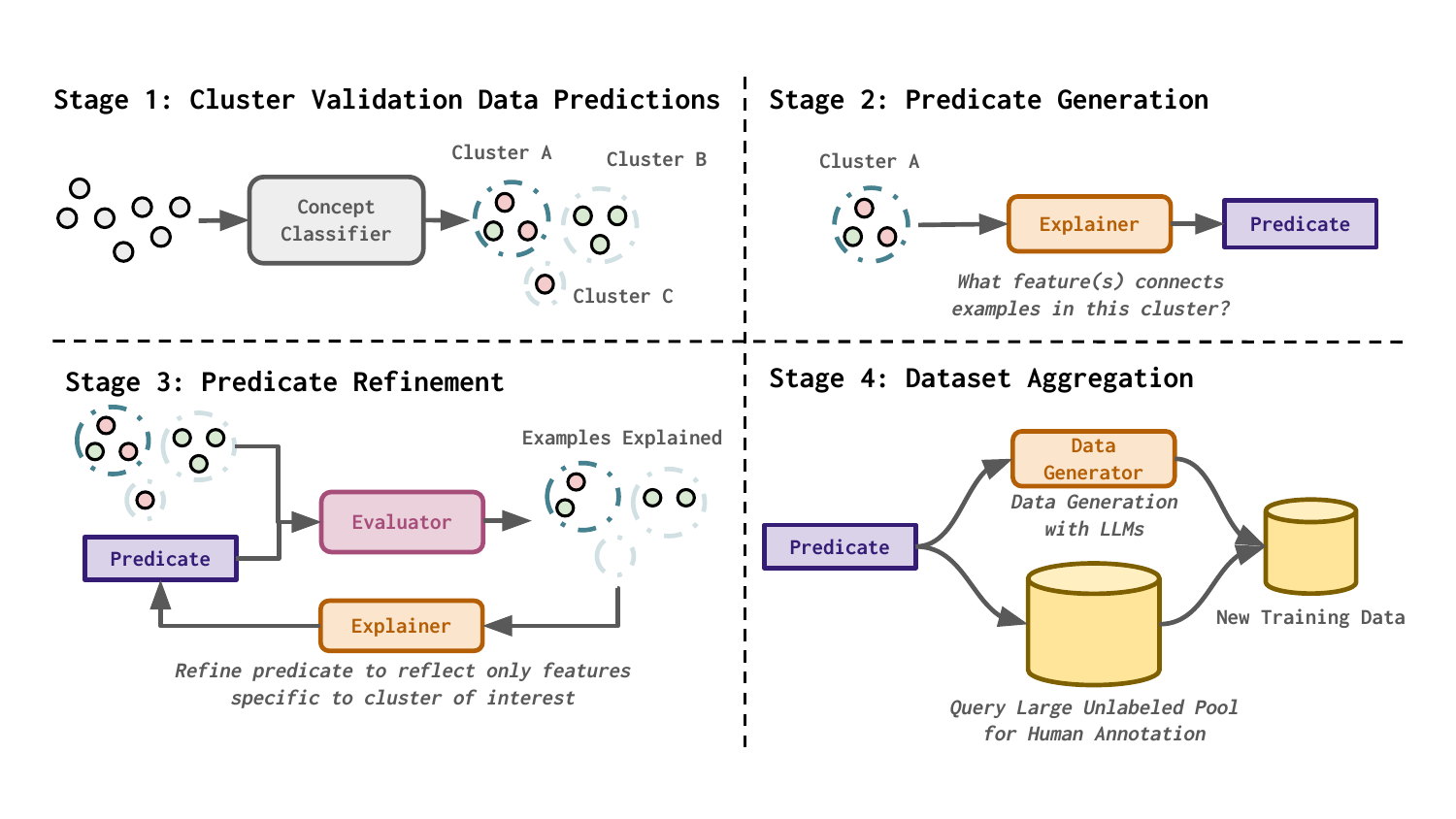}
    % \vspace{-2em}
    \caption{Overview of our classifier debugging framework, \discern. 
    The framework comprises four stages: (1) clustering validation set examples to identify data sub-populations where the classifier makes most errors, (2) cluster description generation using an explainer LLM, (3) refining cluster descriptions through interaction between the explainer and evaluator for higher precision, and (4) model refinement through dataset aggregation.
    }
    \label{fig:intro_framework}
\end{figure*}
Building on recent advancements in large language models \citep[LLMs,][]{achiam2023gpt, touvron2023llama, jiang2023mistral}, 
we aim to bridge this gap with open-ended natural language descriptions of error types.
Such descriptions can offer two major advantages: (1) language descriptions can help structure the generations or acquisitions of new labeled examples, and (2) articulation can allow developers to audit and intervene in the debugging process.
With this premise, we introduce \discern, an iterative approach to improve text-classfiers using \textit{precise} natural language descriptions of their systematic errors
(see Figure \ref{fig:intro_framework}).
\discern utilizes off-the-shelf large language models for distinct roles. An \textbf{explainer LLM} is used to generate predicate-style descriptions\footnote{Predicate-style descriptions refer to concise statements in natural language that describe characteristics or patterns observed within a specific subset of data.} for underperforming clusters of training examples.
To enhance precision, \discern refines the predicates identified by the explainer through an interaction loop.
In this loop, an \textbf{evaluator LLM} assesses whether the predicate applies exclusively to examples within a given cluster. 
Using this feedback on the examples successfully explained by the predicate as well as those that it struggles to explain, the explainer dynamically adjusts its prediction until a desired precision threshold is achieved.
Finally, the generated descriptions are utilized to augment training sets, either through data augmentation using a \textbf{data-generator LLM} or active learning, to retrain and improve the classifier.
In experiments, on a set of three different text-classification tasks, we demonstrate the utility of descriptions generated by our framework in identifying meaningful systematic biases in classifiers.
On the AGNews dataset, by augmenting the training set with synthetically generated instances, we are able to achieve statistically significant improvements over baseline approaches that generate instances from examples of biased instances alone (Section \ref{sec:results_and_analyses}).
In other data sets, \discern can reduce misclassification rates in biased clusters by at least 10\%.
Importantly, we show that language explanations of systematic biases are more helpful for users, and they are 25\% more effective in identifying new biased instances (Table \ref{tab:human_eval}).
Finally, we evaluate the multiple design choices that constitute our framework and ascertain the capacity of our framework to enhance its performance in conjunction with the integration of larger and more robust language models.
\\
\\
Our contributions are as follows:
\begin{itemize}[itemindent=0.3em, labelsep=0.2cm, leftmargin=0.6em, itemsep=0.0em]
    \item A framework for generating precise natural language explanations of systematic errors in models designed for text classification tasks. The precision of the explanations enables a deeper understanding of the underlying biases and aids in developing effective mitigation strategies.
    
    \item Quantitative evaluations demonstrating the value of \discern's explanations through improved classifier performance by synthetic data augmentation and active learning.\footnote{Code to reproduce experiments will be released on first publication.}

    \item Qualitative evaluations of \discern explanations against other approaches emphasize the crucial role of explanations in an efficient and effective understanding of systematic biases.

    \item We analyze the role of different design choices that lead to the generalizability of our framework and outline opportunities for improvement.
\end{itemize}

\section{Related Work}
\label{sec:related_work}
\paragraph{Automatic Failure Discovery.}

To identify failure modes in model predictions, early works employ manual inspection of model prediction errors \citep{vasudevan2022when} or hypothesis testing \citep{poliak-etal-2018-hypothesis}, or adversarial testing \citep{ribeiro-etal-2018-semantically, kiela-etal-2021-dynabench}. 
However, manual inspection requires extensive domain expertise and can be labor intensive.

More recent efforts propose automatic frameworks that approach this problem through the lens of \textit{slice discovery} \citep{eyuboglu2022domino, deim2023hua}, where a \textit{slice} represents a portion of the dataset where the model more frequently makes errors in inference. 
Closer to our work, \citet{rajani-etal-2022-seal} propose SEAL, an interactive visualization tool to describe examples that exhibit high errors using natural language. 
Different from the objective of this work, we propose to use natural language descriptions as an interpretable medium to refine text-based classifiers.

\paragraph{Model Refinement.} To tackle the challenge of underperforming subgroups, previous work has also proposed multiple distributionally robust training strategies \citep{Sagawa*2020Distributionally, pmlr-v139-liu21f,sohoni2020no}. 
Note that these objective functions are complementary to our work and in principle could be utilized to enhance model performance (see \cite{lee2024clarify} for how to use language explanations to perform robust optimization).
However, according to \citet{he-etal-2023-targeted}, these objectives improve the performance of challenging subgroups at the expense of overall accuracy.
We follow the recommendation in \citet{he-etal-2023-targeted} and use data augmentation and active learning to demonstrate the utility of our approach.

\paragraph{LLM Refinement.} LLMs, while adept at many tasks without prior training, struggle with more challenging tasks. 
As a result, recent studies propose to refine LLM predictions through an iterative verification process.
% 
% % 
\textsc{Self-Refine} \citep{madaan2024self} proposes iterative feedback generation and refinement of predictions to enhance performance in text and code generation tasks, while \textsc{Self-Debugging} \citep{chen2024teaching} advocates leveraging unit test execution results to enhance code quality.
In contrast to these studies, we use refinement to understand classifier behavior, not to enhance individual predictions from LLMs.

\paragraph{Data Augmentation with LLMs.} With the growing capabilities of LLMs, recent works have proposed to use LLMs to generate examples to supervise machine learning models \citep{whitehouse-etal-2023-llm, dai2023auggpt}.
Our work differs from these augmentation models in that with \discern we infer the high-level semantic concept that connects existing examples before performing augmentation. 
In other words, natural language (NL) statements act as the intermediate for the augmentation step in our procedure. 
The benefits of the same can be observed throughout our experiments. Our work can be considered as an improvement that can complement methods in \citet{whitehouse-etal-2023-llm} and \citet{dai2023auggpt} that effectively perform example-based augmentation (our No Description baseline).

\section{Method}
\label{sec:method}
In this section, we first formally define our problem setup (\secref{sec:problem_setup}). Next, we provide detailed descriptions of the key stages in \discern (\secref{sec:main_framework}).

\subsection{Problem Setup}
\label{sec:problem_setup}

We consider a classifier denoted as \( f: \mathbb{X} \rightarrow \mathbb{Y} \), where \( \mathbb{X} \) represents textual inputs, such as sentences, and \( \mathbb{Y} \) denotes the corresponding set of classification labels for a specific task (e.g., sentiment analysis). The classifier has been initially trained on a dataset, \( \mathcal{D}_{train} \). However, it is prone to acquiring spurious correlations between the inputs and outputs due to prevalent issues such as annotation artifacts \citep{gururangan-etal-2018-annotation, mccoy-etal-2019-right} or inadequate support for certain classes within the dataset \citep{Sagawa*2020Distributionally}. Our goal, given a validation dataset, \( \mathcal{D}_{val} \), is to identify and describe clusters where the misclassification rate exceeds the classifier’s general misclassification rate.
Formally, we identify clusters $c$ such that, $\mathbb{E}_{(x,y) \sim \mathcal{D}_{val, c}} [f(x) \neq y] > \mathbb{E}_{(x,y) \sim \mathcal{D}_{val}} [f(x) \neq y]$ and utilize the examples in these clusters to inform future classifier refinement.
Rather than directly leveraging these problematic examples to augment the training dataset with additional labeled instances, we demonstrate the value of generating natural-language explanations as an intermediary in the process.
This strategy not only augments the interpretability and understanding of the model, but can efficiently improve classifier performance.

\subsection{\discern}
\label{sec:main_framework}

Broadly, our framework, \discern, is composed of four stages: (1) clustering of validation set examples, (2) predicate generation, (3) predicate refinement, and (4) model refinement with dataset aggregation. Figure \ref{fig:intro_framework} illustrates these four stages in our framework. 

\paragraph{Clustering validation set examples.} In this stage, we target the detection of systematic biases: situations where the model consistently underperforms on data points that exhibit common characteristics or features. Our goal is to elucidate these biases by identifying sub-populations within the data that share similar features. 
For this, we perform agglomerative clustering\footnote{We use the sklearn implementation of AgglomerativeClustering with a distance-based threshold.} 
over the data points in the validation set based on their sentence embeddings. We use the \texttt{text-embedding-3-small} embeddings for clustering, as these embeddings encode semantics of the text, thus ensuring that the systematic biases we identify are substantively grounded in semantic meaning. 
Following clustering, we compute the classifier misclassification rate on different clusters and generate predicates exclusively for those clusters that have a higher misclassification rate than the base misclassification rate of the classifier (\secref{sec:problem_setup}).

\paragraph{Predicate generation.} Using the examples from the clusters that exhibit a high misclassification rate, we prompt an \textbf{explainer LLM}
(in our experiments, \texttt{gpt-3.5-turbo-0125})
to generate descriptions that precisely capture the defining characteristics of the examples within these clusters. 
Drawing on recent work in prompting for planning and reasoning \citep{yao2023react}, we employ thought-based prompting to effectively guide the model to identify and articulate the common characteristics that link examples in a cluster. 
A detailed list of prompts used through different stages of the framework is provided in Appendix \ref{sec:prompt_templates}.

\paragraph{Predicate refinement.} 
The explainer LLM in the previous step is directed to generate descriptions that recall the distinctive characteristics of examples within a cluster.
However, the resultant descriptions often lack \textit{specificity} and encompass examples that belong to multiple clusters. 
In other words, the descriptions do not accurately capture the factors that cause the classifier to perform poorly in a particular cluster, thus inadequately representing the systematic bias.
Past work suggests that augmenting datasets using inadequate strategies can result in a decrease in overall classifier performance \citep{ribeiro-lundberg-2022-adaptive}.
To ensure description specificity for understanding classifier behavior on a target data cluster, 
we need to ensure that the explaining chain can reason over examples within the cluster and those outside it.

To achieve this, we first assess the specificity of the generated descriptions using an evaluator function, which we refer to as the \textbf{evaluator LLM}.
The evaluator LLM, instantiated using a secondary LLM, guides the explainer LLM by identifying examples within the target cluster and outside of it that align with the description generated previously. To evaluate alignment, we prompt the evaluator LLM to check if each example in the target cluster (and outside it) satisfies the predicate in the description.\footnote{The prompt to achieve this can be found Appendix \secref{sec:prompt_templates}.} 
Subsequently, the explainer LLM uses the information of the in-cluster and out-of-cluster examples to refine its description to be more precise. 
We repeat this process until the refined description passes a specific threshold, measured by the evaluator LLM. 
This threshold is based on the percentage of examples that are satisfied within the target cluster versus those outside it by the description generated using the explainer LLM.
Through our iterative refinement process, the model can identify specific characteristics of clusters that explain the systematic bias associated with a classifier. 

\paragraph{Model refinement with dataset aggregation.} Given the descriptions that have been generated for the classifier, we now focus on how to utilize these descriptions to improve the classifier. 
In this work, we adopt two different strategies for improving the classifiers given descriptions: (1) \textbf{synthetic dataset augmentation} -- where we prompt a \textbf{data-generator LLM} using the iteratively refined descriptions to generate new examples for the classification task, 
and (2) \textbf{active learning} -- where we assume access to a pool of unlabeled examples and augment the training set with annotations for examples that match our descriptions.

\section{Experiments}
\label{sec:experiments}
In this section, we outline our experimental procedures to evaluate \discern.

\paragraph{Datasets.}
We use three multiclass text-classification datasets: (1) TREC \citep{li-roth-2002-learning} -- a six-class classification task comprising of questions labeled according to the type of the question, (2) AG News \citep{Zhang2015CharacterlevelCN} -- a collection of news articles labeled according to the category of the article, and (3) COVID Tweets \citep{covidtweets} -- a sentiment-classification task that classifies tweets related to COVID.

\paragraph{Classifiers.} 
Our framework is designed for developers who need to provide low-latency, high-throughput ML solutions with minimal bias. 
For our experiments, we hence use \href{https://huggingface.co/distilbert/distilbert-base-uncased}{\texttt{distilbert-base-uncased}} and \href{https://huggingface.co/FacebookAI/roberta-large}{\texttt{roberta-large}}, as they are sufficient to handle complex tasks while being light enough for mobile deployment. 
This ensures that our approach is practical and effective for real-world applications, enabling users to benefit from advanced ML capabilities on mobile devices offline, without compromising performance or fairness.
Hence, we assume only the developer has access to the LLMs while the user at test-time does not have access to a model as complex. 

These classifiers are initially trained on a subset of the complete training set to simulate realistic learning scenarios with limited data. 
Validation sets are subsampled to match these subsets, and the remaining training data is used as an unlabeled pool for active learning experiments.

\begin{table*}[]
    \centering
    \scalebox{0.99}
    {
    \begin{tabular}{l|cc|cc|cc}
    \toprule
    \textbf{Dataset $\rightarrow$} & \multicolumn{2}{c|}{\textbf{TREC (2000)}} & \multicolumn{2}{c|}{\textbf{AGNews (1500)}} & \multicolumn{2}{c}{\textbf{Covid (4000)}} \\
    \textbf{\# Aug. Ex.} &       \textbf{500} &     \textbf{1000} &          \textbf{500} &     \textbf{1000} &        \textbf{500} &     \textbf{1000} \\
    \midrule
    Base & \multicolumn{2}{c|}{$58.48$} & \multicolumn{2}{c|}{$75.8$} & \multicolumn{2}{c}{$47.68$} \\
    \midrule
    \textbf{No Exp.           } &   $77.09_{(2.18)}$ &  $78.04_{(1.74)}$ &      $80.03_{(1.51)}$ &  $80.68_{(1.08)}$ &    $51.07_{(0.66)}$ &  $48.08_{(1.14)}$ \\
    % \textbf{DEIM*             } &   $71.91_{(4.20)}$ &  $75.69_{(3.14)}$ &      $\mathbf{80.52_{(0.77)}}$ &  $80.76_{(1.33)}$ &    $51.16_{(0.34)}$ &  $\mathbf{49.29_{(0.80)}}$ \\
    \textbf{\discernf             } &   $76.99_{(2.49)}$ &  $78.98_{(1.88)}$ &      $79.75_{(1.25)}$ &  $80.96_{(1.98)}$ &    $51.12_{(0.64)}$ &  $48.60_{(1.36)}$ \\
    \textbf{\textsc{{DiScErN}}} &   $\mathbf{77.20_{(1.80)}}$ &  $\mathbf{79.21_{(1.53)}}$ &      $80.39_{(1.35)}$ &  $\mathbf{83.44_{(1.00)}}^\dagger$ &    $\mathbf{51.55_{(0.47)}}$ &  $49.06_{(0.79)}$ \\
    \bottomrule
    \end{tabular}
    }
    \caption{Accuracy of \texttt{distilbert-base-uncased} classifier after augmenting the training set with examples that have been generated using different approaches. Numbers in brackets next to dataset names indicate the number of training examples used for learning the initial classifier. \textbf{Bold} numbers indicate the best average classifier accuracy across five runs. $^\dagger$ indicates statistically significant improvement over other approaches using t-test. }
    \label{tab:dataset_generation}
\end{table*}

\paragraph{Metric.} We assess the utility of precise and semantically meaningful natural language explanations from \discern by evaluating the performance of classifiers trained with augmented data on the validation set.
We report the average performance and standard deviation across five random seeds, unless otherwise noted.

\paragraph{Baselines.} For the dataset augmentation experiments, we use two baselines to compare with \discern. 
The first baseline is a naive augmentation (or no descriptions) baseline (No Desc.), where we 
generate additional instances that adhere to the style and semantic content of the cluster exemplars.
This baseline helps us to evaluate the role of natural language as a bottleneck for successful model debugging.
To establish the value of refinement, our second baseline uses the explanations generated without iterative refinement to augment the training dataset. 
We refer to this baseline as \discernf\footnote{F for first explanation generated by the explainer LLM.}.
It is worth noting that this baseline, while sharing similarities with the visualization approach proposed in \citet{rajani-etal-2022-seal}, is distinct in its application for classifier improvement.\footnote{This is not an exact replication of SEAL as we use more recent LLMs with the thought-based prompting.}

\paragraph{Experimental Setup.} We utilize \texttt{gpt-3.5-turbo-0125} as our \textbf{explainer LLM} and \textbf{data generator LLM}. 
As our \textbf{evaluator LLM}, we use \href{https://huggingface.co/mistralai/Mixtral-8x7B-Instruct-v0.1}{\texttt{Mixtral-8x7B-Instruct}} \citep{jiang2024mixtral}, a recent open source instruction-tuned large language model.\footnote{We evaluate other choices of predicate evaluators in \secref{sec:ablations}.}
Choosing the evaluator LLM to be different from the explainer LLM, allows us to be leverage the diverse perspectives from different models and avoid confirmation bias \citep{panickssery2024llm}.
This strategic choice also serves as a safeguard against the potential pitfalls of confirmation bias, thus ensuring the quality and accuracy of cluster characterizations.
We set the refinement threshold as recognizing more than 80\% of examples within a target cluster and less than 20\% of the examples outside the cluster and the maximum number of refinement iterations to five.
For fair comparison across methods, we only perform dataset augmentation for those clusters that have passed the refinement threshold.
Additionally, we do not alter training hyperparameters between pre-debugging and post-debugging stages.\\
\\
A full list of hyperparameters used in our experiments can be found in Appendix \secref{sec:training_details}.

\vspace{-0.6em}
\section{Results and Analyses}
\label{sec:results_and_analyses}

\paragraph{Generating synthetic examples using \discern descriptions leads to significant classifier improvement.} 
We evaluate the accuracy of the \texttt{distilbert} classifier, fine-tuned with examples generated by various methods.
Table \ref{tab:dataset_generation} shows the impact of using 500 and 1000 augmented examples on classifier performance.
First, we observe that descriptions of both \discern and \discernf improve over the naive augmentation baseline in most settings, with \discern showing marginal statistical significance (paired t-test; p-value$= 0.05$) across three datasets and augmentation configurations.
This highlights the utility of language descriptions in designing classifier debugging frameworks.
Second, our proposed method, \discern, consistently outperforms \discernf, showing the benefit of high-precision descriptions with marginal statistical significance (paired t-test; p-value$=0.09$).
Furthermore, in the AGNews news classification task, we note that the addition of 1000 synthetic examples leads to a substantial improvement ($\sim3\%$ absolute) in classifier accuracy.\footnote{The accuracy improvements obtained for \discern in this setting is statistically significant compared to naive augmentation and \discernf baselines using an independent samples t-test ($p<0.05$).}

\begin{table*}
    \parbox{4.5cm}{
    \centering
    \scalebox{0.83}{
    \begin{tabular}{l|c|c}
    \toprule
    \textbf{Method} & {\textbf{TREC}} & {\textbf{Covid}} \\
    \midrule
    Base &              $100.00$ &                 $72.73$ \\
    \midrule
    No Desc. &                 $3.17$ &                 $30.95$ \\
    \discernf &                 $4.76$ &                 $40.91$ \\
    \discern &                 $0.00$ &                 $27.78$ \\
    \bottomrule
    \end{tabular}
    }
    \caption{Median misclassification rates for erroneous clusters before (Base) and after training of a \texttt{distilbert} classifier with 1000 augmented examples using different approaches. 
    % \ssr{I don't see a pre or post training mention in the table. How do I read this table?}
    }
    \label{tab:dgen_misclassify}
    }
    \qquad
    \parbox{10.5cm}{
    \centering
    \scalebox{0.85}{
    \begin{tabular}{l|cc|cc}
    \toprule
    \textbf{Dataset $\rightarrow$} & \multicolumn{2}{c|}{\textbf{TREC (1500)}} & \multicolumn{2}{c}{\textbf{AG News (500)}} \\ % & \multicolumn{2}{c}{\textbf{Covid (4000)}} \\
    \textbf{\# Aug. Ex.} &       \textbf{500} &     \textbf{1000} &         \textbf{500} &      \textbf{1000} \\% &        \textbf{500} &     \textbf{1000} \\
    \midrule
    Base & \multicolumn{2}{c|}{$70.85$} & \multicolumn{2}{c}{$41.2$}\\% & \multicolumn{2}{c}{$55.1$} \\
    \midrule
    No Desc. &  $71.99_{(14.04)}$ &  $85.26_{(3.13)}$ &    $\mathbf{58.88_{(10.57)}}$ &   $61.96_{(9.47)}$\\% &    $50.76_{(1.42)}$ &  $46.45_{(3.07)}$ \\
    \discernf &  $72.25_{(11.95)}$ &  $86.51_{(1.98)}$ &    $58.60_{(10.35)}$ &  $64.28_{(10.67)}$\\% &    $\mathbf{52.35_{(1.48)}}$ &  $\mathbf{48.08_{(1.39)}}$ \\
    \textsc{\discern} &   $\mathbf{78.11_{(3.38)}}$ &  $\mathbf{88.54_{(0.90)}}$ &    $55.44_{(14.50)}$ &  $\mathbf{67.00_{(10.39)}}$\\% &    $51.75_{(1.76)}$ &  $47.64_{(0.91)}$ \\
    \bottomrule
    \end{tabular}
    }
    \caption{Accuracy of \texttt{roberta-large} classifier after augmenting the training set with examples that have been generated using different approaches. Numbers in brackets next to the names of the dataset indicate the number of training examples used to learn the initial classifier. \textbf{Bold} numbers indicate the best average classifier accuracy across five runs. 
    Full results in Table \ref{tab:dataset_generation_roberta_appendix}.
    % $^\dagger$ indicates statistically significant improvement over other approaches using t-test.
    }
    \label{tab:dataset_generation_roberta}
    }
\end{table*}

In Table \ref{tab:dgen_misclassify}, we show that 
\discern substantially reduces misclassification rates in the underperforming clusters on the other two datasets. 
Specifically, for the TREC and Covid datasets, \discern achieves perfect classification and reduces the misclassification rate to 27.78\%, respectively.
Compared to the baseline of naive augmentation (No Desc.), we observe that \discern has a substantial improvement in misclassification rates.
More notably, we observe the value of \textit{precision} in language descriptions by comparing the result with \discernf, whose misclassification rates are worse than the naive augmentation baseline.

\begin{figure}[h!]
    \centering
    \includegraphics[width=0.99\linewidth]{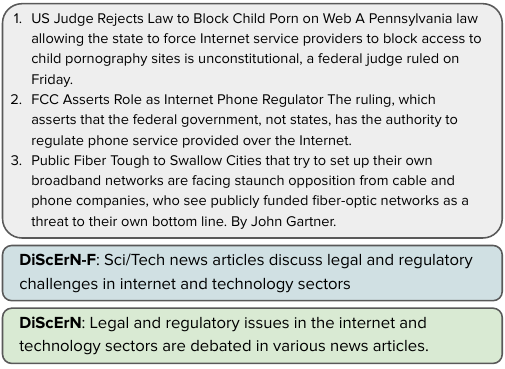}
    \caption{Example of descriptions generated by \discern and \discernf for an underperforming cluster in the AGNews dataset. Examples for descriptions with other datasets can be found in the Appendix.}
    \label{fig:single_example}
    % \vspace{-1.5em}
\end{figure}

Figure \ref{fig:single_example} presents descriptions generated by \discernf and \discern for the AGNews datasets (examples for other datasets in Figure \ref{fig:examples} in the Appendix).
From the descriptions, we can observe the ability of \discern to capture the nuances that enable targeted improvement.
In particular, \discern descriptions provide a more precise observation of \textit{``debate"} in the corresponding news articles, as opposed to \discernf. 
Put together, these findings underscore the potential of \discern to improve classifiers by addressing systematic errors.

\begin{figure*}[t!]
    \centering
    \includegraphics[width=0.99\textwidth]{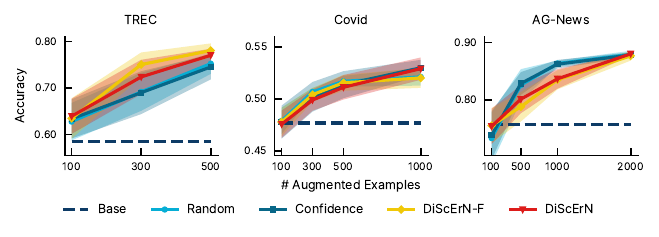}
    % \vspace{-1.5em}
    \caption{Average accuracy of \texttt{distilbert-base-uncased} classifiers after augmenting the training set with examples identified and annotated from a large unlabeled pool using different approaches. Shaded regions indicate the standard deviation over five runs.
    }
    \label{fig:active-learning}
\end{figure*}
\paragraph{\discern improvements generalize across models.} 
We evaluate the performance of a different classifier model, \texttt{roberta-large}, to assess the generalization of the observed improvements.
In Table \ref{tab:dataset_generation_roberta}, we compute the accuracy of the classifier following the augmentation of the training set with examples generated through different approaches.
Similar to the results for the \texttt{distilbert} classifier, we observe consistent improvements in classifier accuracy using examples generated using \discern descriptions as opposed to the no-description baseline. 
This highlights the classifier-agnostic utility of our framework in identifying systematic errors and rectifying them through data augmentation.

\paragraph{Active Learning using \discern Descriptions.}

Language descriptions derived using our method can also be used to identify examples from an unlabeled pool that could help improve classifier performance.
Consequently, we employ this strategy to identify examples from the unlabeled pool of each of these datasets.
Specifically, given the language descriptions, from \discern and \discernf, we use the \texttt{Mixtral-8x7B-Instruct} model to identify examples that satisfy the predicate mentioned in the description.
All examples identified through this process are then added to the training set to retrain the classifier.
We measure the classifier accuracies post training with new training set.
We use two standard active learning baselines: (a) \textit{random} -- annotating and augmenting random examples from the unlabeled pool, and (b) \textit{confidence} -- selecting examples predicted with least classifier confidence for annotation and augmentation.
Recent work indicates these strategies remain competitive for active learning with large language models \citep{margatina-etal-2023-active}.

In Figure \ref{fig:active-learning}, we plot the accuracy of the classifier as a function of the number of annotated examples incorporated into the training process.\footnote{The x-axis varies based on the size of the unlabeled pool for each dataset and the number of labeled examples identified by the description-based methods.}
We make a couple of observations. 
First, the addition of examples suggested by \discern is better than the addition of random samples to the training set, especially on the TREC and Covid datasets.
This suggests that \discern is adept at identifying informative examples to improve the classifier.
Second, the confidence-based approach predominantly outperforms description-based approaches, particularly when few examples are added to the training set.
However, it is noteworthy that the improvement achieved through the \discern suggested examples gradually catches up as the number of annotated examples increases.

\begin{table*}[]
    \centering
    \scalebox{0.8}
    {
    \begin{tabular}{l|cc|cc|cc}
    \toprule
    \textbf{Dataset $\rightarrow$} & \multicolumn{2}{c|}{\textbf{TREC (2000)}} & \multicolumn{2}{c|}{\textbf{AGNews (1500)}} & \multicolumn{2}{c}{\textbf{Covid (4000)}} \\
    \textbf{\# Aug. Ex.} &       \textbf{500} &     \textbf{1000} &          \textbf{500} &     \textbf{1000} &        \textbf{500} &     \textbf{1000} \\
    \midrule
    Base & \multicolumn{2}{c|}{$58.48$} & \multicolumn{2}{c|}{$75.8$} & \multicolumn{2}{c}{$47.68$} \\
    \midrule
    \textbf{No Exp.           } &   $77.09_{(2.18)}$ &  $78.04_{(1.74)}$ &      $80.03_{(1.51)}$ &  $80.68_{(1.08)}$ &    $51.07_{(0.66)}$ &  $48.08_{(1.14)}$ \\
    \textbf{DEIM*             } &   $71.91_{(4.20)}$ &  $75.69_{(3.14)}$ &      $\mathbf{80.52_{(0.77)}}$ &  $80.76_{(1.33)}$ &    $51.16_{(0.34)}$ &  $\mathbf{49.29_{(0.80)}}$ \\
    \textbf{\discernf             } &   $76.99_{(2.49)}$ &  $78.98_{(1.88)}$ &      $79.75_{(1.25)}$ &  $80.96_{(1.98)}$ &    $51.12_{(0.64)}$ &  $48.60_{(1.36)}$ \\
    \textbf{\textsc{{DiScErN}}} &   $\mathbf{77.20_{(1.80)}}$ &  $\mathbf{79.21_{(1.53)}}$ &      $80.39_{(1.35)}$ &  $\mathbf{83.44_{(1.00)}}^\dagger$ &    $\mathbf{51.55_{(0.47)}}$ &  $49.06_{(0.79)}$ \\
    \bottomrule
    \end{tabular}
    }
    \caption{Accuracy of \texttt{distilbert-base-uncased} classifier after augmenting the training set with examples that have been generated using different approaches. Numbers in brackets next to dataset names indicate the number of training examples used for learning the initial classifier. \textbf{Bold} numbers indicate the best average classifier accuracy across five runs. $^\dagger$ indicates statistically significant improvement over other approaches using t-test. }
    \label{tab:dataset_generation_appendix}
\end{table*}

\paragraph{\discern outperform keyword-based approaches.} Prior work in NLP has proposed to identify clusters using manually prescribed keywords, typically provided by domain experts \cite{deim2023hua}. 
Here, we compare the performance of \discern against the approach \cite{deim2023hua}. 
To this end, we utilize \texttt{gpt-3.5-turbo} to generate keywords describing the semantic content for the datasets we use in our experiments.
Next, we assign clusters (obtained from agglomerative clustering) towards one of the keywords.
Finally, we use the keywords for the cluster to guide the generation for new examples and re-train the classifier.
This approach roughly follows the work in DEIM \cite{deim2023hua} and hence we name it DEIM*.
In Table \ref{tab:dataset_generation_appendix}, we present the results from the comparison.
On average, we observe that \discern beats the keyword-based approach with statistical significance in the AGNews dataset.
This further underscores the advantage of free-form language descriptions of underperforming clusters.

\paragraph{Language descriptions facilitate a more effective and efficient understanding of biases among users.} 

Here we explore how language descriptions help users understand biases in classifiers and identify likely misclassified instances. 
This concept aligns with simulatability from previous explainability research \citep{hase-bansal-2020-evaluating, menon-etal-2023-mantle}, which assesses users' comprehension of classifier predictions.

To do this, we conduct a user study, in which users are shown examples or \discern descriptions of clusters where the classifier has a higher misclassification rate than its base rate. 
Based on the information provided, users are tasked with identifying if new examples, drawn from one to two erroneous clusters per dataset, match the characteristics of given descriptions or cluster exemplars.
The test uses new examples from within and outside the erroneous clusters, the latter having a high BERTScore \citep{Zhang*2020BERTScore:} similarity with at least one example in the cluster.
Participants provided predictions for six new examples in each HIT, and we measured the accuracy of predicting examples belonging to the erroneous cluster.
24 workers took part in this study conducted on Prolific and were compensated at $\$12$/hr.

\begin{table}[h]
    \centering
    \scalebox{0.99}{
    \begin{tabular}{l|r|r|r}
    \toprule
    \textbf{Method} & \textbf{Acc. ($\uparrow$)} & \textbf{Time ($\downarrow$)} & \textbf{Help. ($\uparrow$)} \\ 
    \midrule
    No Desc. & $62.5$\% & $185$s & $3.00$\\
    \discern & $\mathbf{79.2\%}$ & $\mathbf{177}$\textbf{s} & $\mathbf{3.83^*}$\\
    \bottomrule
    \end{tabular}
    }
    \caption{User evaluations in understanding classifier biases based on cluster exemplars (No Desc.) vs \discern descriptions.* = marginal statistical significance with t-test (p-value $< 0.1$). }
    \label{tab:human_eval}
    % \vspace{-1.5em}
\end{table}

In our results (Table \ref{tab:human_eval}), 
show that after reviewing \discern descriptions, users accurately predict new examples that exhibit characteristics similar to those in the erroneous clusters in 79.2\% of cases, compared to 62.5\% without descriptions.

Further, users provided with descriptions required less time to perform the task and found them more helpful. 
These findings underscore the potential of language descriptions in enhancing users' understanding of systematic biases in classification models, a crucial step towards designing fairer and equitable models for real-world deployment.

\section{Ablations}
\label{sec:ablations}

\paragraph{Impact of Embeddings used during Clustering.}
We examine the impact of different embeddings on the initial stage of the \discern framework, specifically focusing on clustering datapoints in the validation set. 
We compare the OpenAI embeddings Ada and v3 here
for the TREC dataset.
As shown in Table \ref{tab:clustering_ablation}, across all methods, utilizing the v3 embedding consistently yields higher accuracy compared to Ada.
This finding underscores the importance of choosing effective embeddings for identification of biases using our framework.
Additionally, we observe that \discern outperforms the naive augmentation (No Descriptions) baseline even when employing weaker embeddings, highlighting the versatility of our framework. 
\begin{table}[h!] % {r}{7.5cm}
    \centering 
    \scalebox{0.99}{
    \begin{tabular}{l|c|c}
    \toprule
    \textbf{Method} &         Ada & v3 \\
    \midrule
    No Descriptions &            $65.76$ &   $\mathbf{78.04}$\\
    \discernf &            $67.61$ &   $\mathbf{78.98}$ \\
    \discern &            $68.18$ &   $\mathbf{79.21}$ \\
    \bottomrule
    \end{tabular}
    }
    \caption{Classiifer accuracies post synthetic data augmentation using different embeddings to cluster validation set datapoints on the TREC dataset.}
    \label{tab:clustering_ablation}
    % \vspace{-1em}
\end{table}

\paragraph{Stronger Explainers enhance Classifier Performance.}
Table \ref{tab:descriptor_ablation} evaluates the impact of different language models used for describing underperforming clusters and their subsequent classifier improvement.
Specifically, it compares the accuracy of a \texttt{distilbert} classifier trained using cluster descriptors derived from two distinct language models: \texttt{gpt-3.5-turbo-0125} and \texttt{gpt-4-0125-preview}. 
We compute classifier improvements by adding 1000 synthetic instances generated using the different descriptions.
Both \discernf and \discern show marked improvements across datasets, highlighting the potential for improvement of our approach with larger and more capable language models.
\begin{table}[h!]
    \centering
    \scalebox{0.9}{
    \begin{tabular}{l|c|c}
    % \toprule
    \multicolumn{3}{c}{\begin{tabular}{@{} c @{}}\texttt{distilbert} accuracy with explainer LLM\\changing from \texttt{gpt-3.5-turbo} $\rightarrow$ \texttt{gpt-4-turbo}\end{tabular}}\\
    \toprule
    \textbf{Method} &         \textbf{TREC} &      \textbf{AG News} \\
    \midrule
    Base & $58.48$ & $75.8$ \\
    \midrule
    \discernf &  $78.98 \rightarrow 81.12$ &  $80.96 \rightarrow 86.44$ \\
    \discern &  $79.21 \rightarrow 79.62$ &  $83.44 \rightarrow 86.85$ \\
    \bottomrule
    \end{tabular}
    }
    \caption{Accuracies post synthetic data augmentation using different language models for describing clusters. 
    }
    \label{tab:descriptor_ablation}
\end{table}
\vspace{-1em}
\paragraph{Predicate Evaluators.}
\begin{figure}
    \centering
    \includegraphics[width=0.95\linewidth]{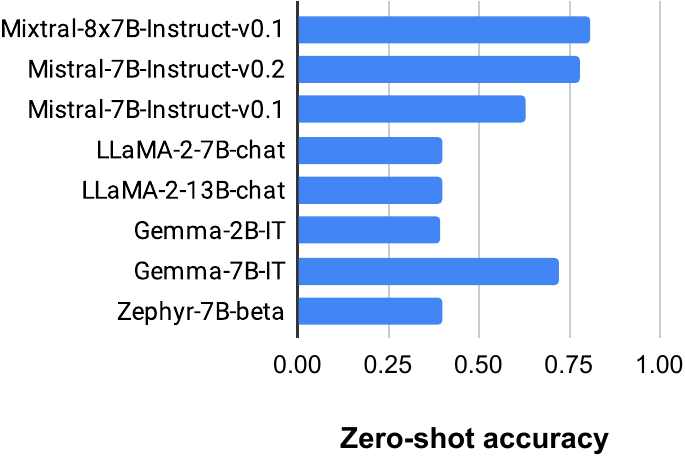}
    % \vspace{-2em}
    \caption{Zero-shot performance of different language models used as predicate evaluators for our task.}
    \label{fig:pred_eval_performance}
    % \vspace{-1em}
\end{figure}

The predicate evaluator in our framework provides signal to the explainer about the alignment of the generated explanations with the examples.
% inside and outside the cluster.
% 
Hence, we need the predicate evaluator to accurately predict whether a predicate applies to a given example.
For this evaluation, we sampled 10 clusters from our three datasets, along with a random collection of datapoints from within and outside these clusters.
We obtain the ``ground-truth'' annotations for the alignment between an explanation and a datapoint using \texttt{gpt-4-turbo} \citep{achiam2023gpt}\footnote{GPT-4 judgments, found to align well with human judgments, serve as our proxy \citep{rafailov2024direct}.}.
In Figure \ref{fig:pred_eval_performance}, we evaluate the performance of different open-source LLMs for this task.
We observe that \texttt{Mixtral-8x7B-Instruct} has the highest agreement with the ground-truth. Consequently, we used it for evaluations in our experiments.

\section{Conclusion}
\label{sec:conclusion}
In this work, we propose a framework \discern, to address systematic biases and improve the performance of text classifiers. 
Using large language models to generate precise natural language descriptions of errors, \discern surpasses example-based augmentation techniques to identify and rectify systematic biases in multiple classifiers across diverse datasets.
Through extensive experimentation, we have demonstrated the capability of \discern in reducing misclassification rates and improving classifier accuracy, consistently outperforming alternative approaches. 
Further, our human evaluations indicates user preference for understanding bias using natural language descriptions.
Overall, our findings underscore the potential of \discern as a powerful tool to improve the performance of text classifiers, thus enabling the design of more reliable and equitable machine learning systems in various domains.
Building on our results, future research directions can explore ways to enhance other applications using the refinement approach used in our work, integrate \discern into informing training recipes (such as, large language model training), and investigate biases transferred between different classifiers.

\section*{Limitations}
\label{sec:limitations}
The exact instantiation of our framework in this work makes use of proprietary large language models. 
The accessibility of these models is contingent upon evolving corporate policies of the respective entities. 
Nevertheless, we believe that with the increasing capabilities of smaller open-source large language models such as \texttt{Mixtral}, we should be able to achieve very similar performance with newer models while being accessible to everyone.

Our agglomerative clustering approach also depends on the distance threshold hyperparameter which can affect the granularity of the explanations.
Future work can look into top-down approaches that can explain classifier biases at the right levels of granularity, thereby enabling interpretability and advancing the efficacy of our framework.

\section*{Acknowledgements}
We would like to thank Kerem Zaman, Jack Goldsmith, Anika Sharma, and the anonymous reviewers for feedback and suggestions on the draft. This work was supported in part by NSF grant DRL2112635. The views contained in this article are those of the authors and not of the funding agency.

\bibliography{custom}
\bibliographystyle{arr_latex/acl_natbib}

\appendix

\section*{Appendix}

In the Appendix, we provide details regarding the compute and training hyperparameters for our experiments (Section \ref{sec:training_details}), prompt templates (Section \ref{sec:prompt_templates}), discuss extended related work (Section \ref{sec:ext_related_work}), provide additional analysis (Section \ref{sec:additional_analysis}), and show example templates used in human evaluation (Section \ref{sec:human_eval_templates}).

\section{Training Details}
\label{sec:training_details}
In Table \ref{tab:classifier_hyperparams}, we detail the hyperparameters used for fine-tuning the \texttt{distilbert-base-uncased} and \texttt{roberta-large} models. 
We maintain the same hyperparameters during re-training of the model using the augmented training set.
We implement these classifiers in Pytorch \cite{NEURIPS2019_9015} using the Huggingface library \cite{wolf-etal-2020-transformers}. 
Classifiers were fine-tuned with full precision on a single NVIDIA A100-PCIE-40GB GPU, 400GB RAM, and 40 CPU cores.

In Table \ref{tab:refine_hyperparams}, we report the hyperparameters for the cluster description generation and synthetic data augmentation.  
Using the agglomerative clustering algorithm and our preset distance thresholds, the number of examples in a cluster typically varies between 10-60.
We use the \href{https://platform.openai.com}{OpenAI API} to make calls to the GPT-3.5 and GPT-4 models. 
We also load the \texttt{Mixtral} model with 16-bit precision (bfloat16). 
The same system configuration, as used for classifier training, is used for these experiments.

In Algorithm \ref{alg:debug_framework}, we summarize the explain and refine iterative setup used in \discern.\\
\\
\textbf{Note:} Since the submission of this work and its eventual acceptance, the codebase for the Mistral tokenizer has been \href{https://huggingface.co/mistralai/Mixtral-8x7B-Instruct-v0.1/discussions/229}{\textbf{modified}} in a way that irreversibly affects its functionality leading to slight differences in numbers. 

\begin{algorithm}
  \caption{\discern}\label{alg:debug_framework}
  \begin{algorithmic}[1]
  \Require Explainer LLM $\mathcal{E}$
  \Require Evaluator LLM $\textsc{P}$
  \Require Data Generator LLM $\mathcal{D_G}$
  \Require validation dataset $\mathcal{D}_{val} = (X_{val}, Y_{val})$
  \Require classifier $f$

  \State // \textbf{Get validation set predictions}
  \State $Y_{pred} = f(X_{val})$
  \State $\mathcal{Y} = \textproc{Unique}(Y_{val})$
  \State // \textbf{Cluster $\mathcal{D}_{val}$ for each class}
  \State $X_{val, y} \leftarrow \{x : (x, y') \in \mathcal{D}_{val}, y = y'\}, \forall y \in \mathcal{Y}$ // \text{Split dataset based on ground-truth label}
  \State $\mathcal{C}_{1:m, y} \leftarrow \textproc{AgglClustering}(X_{val,y}), \forall y \in \mathcal{Y}$
  
  \For {$c \in \mathcal{C}_{1:m, y}$}
    \State iterations = 0
    \State $c_{out} = \{\}$
    \While {\textbf{not} refinement\_threshold\_met \textbf{or} iterations $<$ max\_iterations}
      \If {iterations $>$ 0}
        \State $c_{out} = \textproc{SampleExamples}(\mathcal{C}_{1:m, y} - \{c\})$ // \text{sample out of cluster examples}
      \EndIf
      \State ${e}_{c} \leftarrow \mathcal{E}(c, c_{out})$
      \State // in-cluster evaluation
      \State $r_{in\_cluster} = \textsc{P}(c, {e}_{c})$
      \State // out-of-cluster evaluation
      \State $r_{out\_cluster} = \textsc{P}(\mathcal{C}_{1:m, y} - \{c\}, {e}_{c})$
      \State refinement\_threshold\_met = ($r_{in\_cluster} >$ in-cluster threshold) \textbf{and} ($r_{out\_cluster} < $ out-cluster threshold)
      \State iterations = iterations + 1
    \EndWhile
    \State // Generate data using the explanation
    \State $X', y' \leftarrow \mathcal{D_G}({e}_{c}, c)$
    \State $X_{train}, y_{train} \text{.append}(X', y')$
  \EndFor
  \State \textproc{ReTrainClassifier}($f$, $X_{train}, Y_{train}$)
  \end{algorithmic}
\end{algorithm}

\begin{table}
    \centering
    \begin{tabular}{l|r}
    \toprule
    \textbf{Hyperparameters} & \textbf{Values} \\
    \midrule
    train\_batch\_size &  32 \\
    eval\_batch\_size &  \begin{tabular}{r@{}}
        512 (\texttt{distilbert})\\ 64 (\texttt{roberta}) \\
    \end{tabular}\\ \\
    gradient\_acc.\_steps & 1 \\
    learning\_rate &   \begin{tabular}{r@{}}
        1e-5 (\texttt{distilbert})\\ 2e-5 (\texttt{roberta}) \\
    \end{tabular}\\
    weight\_decay &  0.01 \\
    adam\_beta1 &  0.9 \\
    adam\_beta2 &  0.999 \\
    adam\_epsilon &  0.0 \\
    max\_grad\_norm &  1.0 \\
    num\_train\_epochs &  3 \\
    lr\_scheduler\_type &   linear \\
    warmup\_ratio &  0.0 \\
    warmup\_steps &  600 \\
    seed & 42 \\
    % load\_best\_model\_at\_end & True \\
    optim &    adamw\_torch \\
    \bottomrule
    \end{tabular}
    \caption{Hyperparameters used for fine-tuning pre-trained models used across different datasets.}
    \label{tab:classifier_hyperparams}
\end{table}

\begin{table}
    \centering
    \scalebox{0.8}{
    \begin{tabular}{l|r}
    \toprule
    \textbf{Hyperparameters} & \textbf{Values} \\
    \midrule
    Clustering Alg. & Agglomerative Clustering\\
    Distance Threshold & \begin{tabular}{r@{}}2 (\texttt{openai-v3})\\1.2 (\texttt{openai-ada})\end{tabular}\\
    Explainer LLM & \texttt{gpt-3.5-turbo-0125} \\
    Explainer Temperature & 0.1 \\
    Explainer top-p & 1\\
    \begin{tabular}{@{}l}Max explainer\\ generation tokens\end{tabular} & 512 \\
    \begin{tabular}{@{}l}In-cluster\\ description threshold\end{tabular} & 0.8 \\
    \begin{tabular}{@{}l}Out-of-cluster\\description threshold\end{tabular} & 0.2\\
    \begin{tabular}{@{}l}Num. In-cluster\\ Examples in Prompt\end{tabular} & 64 \\
\begin{tabular}{@{}l}Num. Out-of-cluster\\ Examples in Prompt\end{tabular} & 32 \\
    Evaluator LLM & \begin{tabular}{r@{}}\texttt{Mixtral-8x7B}\\\texttt{Instruct-v0.1}\end{tabular}\\
    \begin{tabular}{@{}l}Max evaluator\\ generation tokens\end{tabular} & 1 \\
    Evaluator precision & bfloat16\\
    Data generator LLM & \texttt{gpt-3.5-turbo-0125} \\
    Generator Temperature & 0.7\\
    Generator top-p & 1\\
    Generator seed & 0\\
    Max generator tokens & 4096 \\
    \begin{tabular}{@{}l}Max generated examples\\(per cluster)\end{tabular} & 100 \\
    \bottomrule
    \end{tabular}
    }
    \caption{Hyperparameters used for generation and refinement of cluster descriptions + synthetic data augmentation.}
    \label{tab:refine_hyperparams}
\end{table}

\begin{figure*}
    \centering
    \includegraphics[width=\linewidth]{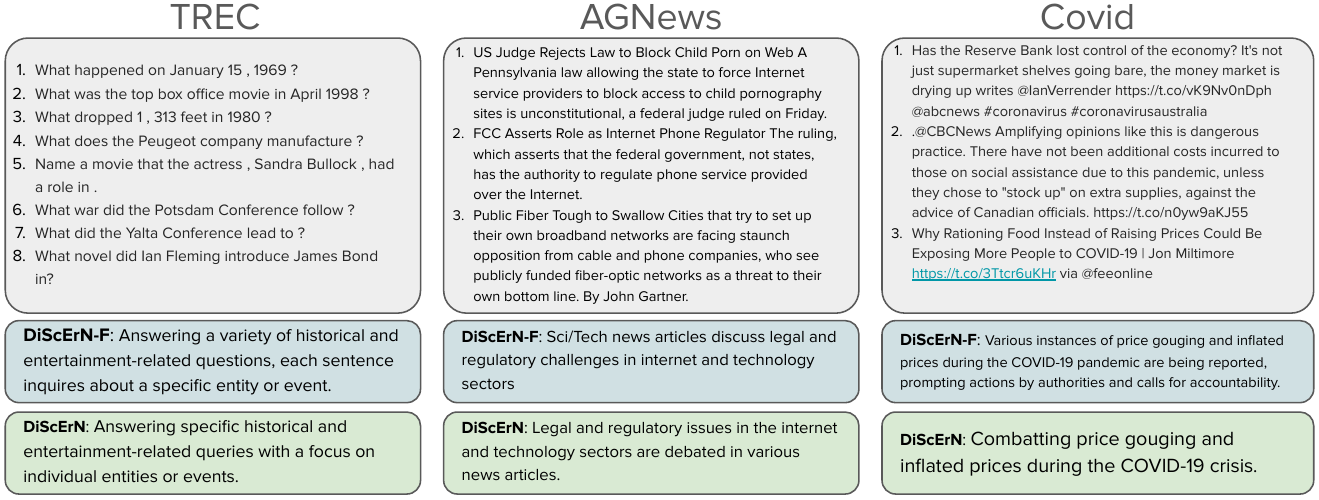}
    \caption{Descriptions generated using \discernf and \discern for erroneous clusters in different datasets using the \texttt{distilbert-base-uncased} classifier.}
    \label{fig:examples}
\end{figure*}

\section{Prompt Templates}
\label{sec:prompt_templates}
In this section, we present the prompt templates used during the different stages of our framework. The table below provides a legend to the exact prompts used for each scenario.
\begin{table}
    \centering
    \begin{tabular}{l|c}
    \toprule
    \textbf{Objective} & \textbf{Reference}\\
    \midrule
        Predicate Generation & Table \ref{tab:prompt_first} \\
        Predicate Refinement & Table \ref{tab:prompt_refine} \\
        Example Evaluation & Table \ref{tab:prompt_align} \\
        Data Generation - Examples & Table \ref{tab:prompt_example_aug}\\
        Data Generation - Explanations & Table \ref{tab:prompt_explanations_aug}\\
    \bottomrule
    \end{tabular}
    \caption{Legend for prompts used in the various stages of \discern.}
    \label{tab:prompt_legend}
\end{table}

\begin{table*}
\centering
\texttt{
\begin{tabular}{|p{15cm}|}
\hline
Here are a group of sentences:\\
\\
\{samples\_in\_prompt\}\\
\\
Generate a single-line predicate description that incorporates the specific word or label `\{label\}'.\\
\\
Your response should be formatted in the following manner:\\
Thoughts:\\
1. The sentences are mainly $<$type of sentences$>$.\\
2. The sentences talk about $<$topic$>$.\\
3. I will also focus on the following attributes about the sentences in the generated predicate to be precise: $<$list of attributes$>$\\
PREDICATE:\\
- ``$<$predicate$>$"\\
\\
Try to make sure that the generated predicate is precise and will only satisfy the examples mentioned above.\\
\\
Thoughts:\\
\hline
\end{tabular}
}
\caption{Prompt used to elicit the first set of explanations given cluster examples alone.}
\label{tab:prompt_first}
\end{table*}

\begin{table*}
\centering
\texttt{
\begin{tabular}{|p{15cm}|}
\hline
You were asked to provide a single-line predicate description for a set of examples (let's call this CLUSTER\_1) shown below:\\
\\
\{samples\_in\_prompt\}\\
\\
You generated the following description: ``\{description\}"\\
\\
This description satisfied the following examples:\\
\\
\{in\_cluster\_satisfied\_examples\}\\
\\
However, the description also identifies with the following examples (that it should not ideally) (let's call this CLUSTER\_2 examples):\\
\\
\{out\_of\_cluster\_satisfied\_examples\}\\
\\
In other words, the current description explains \{pass\_rate:.1f\}\\% examples in CLUSTER\_1 and \{fail\_rate:.1f\}\\% examples in CLUSTER\_2.\\
\\
Please re-write the description that explain only examples from CLUSTER\_1 while excluding examples from CLUSTER\_2.\\
\\
Try to make descriptions simple and general. For example, you could focus on the syntax, topic, writing style, etc.\\
First, for the failing description above, explain why the description does not accomplish the goal of describing only the examlpes in CLUSTER\_1. Output this reasoning as:\\
Thoughts:\\
1. The examples in CLUSTER\_1 and CLUSTER\_2 talk about one common topic: \{label\}.\\
2. The examples in CLUSTER\_1 emphasize on $<$CLUSTER\_1 description$>$.\\
3. Whereas, the examples in CLUSTER\_2 emphasize on $<$CLUSTER\_2 description$>$.\\
4. The previous description failed because $<$reason$>$.\\
5. The examples in CLUSTER\_2 are about "$<$reason$>$" which is not present in CLUSTER\_1. I will focus on mentioning this reason in the new predicate.\\
Then output the description so that it explains only examples in CLUSTER\_1, using the following format:\\
NEW PREDICATE:\\
- ``$<$more precise-yet-simple CLUSTER\_1 description that highlights difference with CLUSTER\_2$>$"\\
\\
Note: The new predicate has to be strictly different from the previous one.\\
Note: Do not mention the words CLUSTER\_1 or CLUSTER\_2 in your new predicate. It should be part of your thought process however.\\
\\
\\
Thoughts:\\
1. The examples in CLUSTER\_1 and CLUSTER\_2 talk about one common topic: \{label\}.\\
\hline
\end{tabular}
}
\caption{Prompt used to refine explanations given in-cluster examples and out-of-cluster examples.}
\label{tab:prompt_refine}
\end{table*}

\begin{table*}
\centering
\texttt{
\begin{tabular}{|p{15cm}|}
\hline
Check if this statement `\{example\}' satisfies the given condition: `\{description\}'. Provide only `Yes' or `No'. When unsure, respond with `No'.\\
\hline
\end{tabular}
}
\caption{Prompt used to check alignment of example with the generated description.}
\label{tab:prompt_align}
\end{table*}

\begin{table*}
\centering
\texttt{
\begin{tabular}{|p{15cm}|}
\hline
In this task, you will be shown some examples sentences that share some property. Your task is to generate 100 more diverse examples that satisfy the shared property of these texts.\\
\\
The examples you generate should follow the style and content of the examples mentioned below:\\
$\{$list\_of\_examples$\}$\\
\\
Consider the linguistic style, content, length, and overall structure of the provided examples. Your generated examples should resemble the provided set in terms of these aspects. Aim to produce sentences that convey similar information or ideas while maintaining consistency in tone, vocabulary, and grammatical structure.\\
\\
Feel free to vary the details and specifics while ensuring that the generated examples capture the essence of the provided set. Pay attention to context, coherence, and any relevant patterns present in the examples to produce outputs that closely align with the given set.\\
\\
Your response:\\
- \\
\hline
\end{tabular}
}
\caption{Prompt used to generate synthetic instances for the classification task using only cluster exemplars.}
\label{tab:prompt_example_aug}
\end{table*}

\begin{table*}
\centering
\texttt{
\begin{tabular}{|p{15cm}|}
\hline
In this task, you will be shown some examples sentences that share a property given by the predicate below. Your task is to generate 100 more diverse examples that satisfy the predicate.\\
\\
Predicate: $\{$predicate$\}$\\
\\
The examples you generate should follow the style and content of the examples mentioned below:\\
$\{$list\_of\_examples$\}$\\
\\
Consider the linguistic style, content, length, and overall structure of the provided examples. Your generated examples should resemble the provided set in terms of these aspects. Aim to produce sentences that convey similar information or ideas while maintaining consistency in tone, vocabulary, and grammatical structure.\\
\\
Feel free to vary the details and specifics while ensuring that the generated examples capture the essence of the provided set. Pay attention to context, coherence, and any relevant patterns present in the examples to produce outputs that closely align with the given set.\\
\\
Your response:\\
- \\
\hline
\end{tabular}
}
\caption{Prompt used to generate synthetic instances for the classification task using descriptions.}
\label{tab:prompt_explanations_aug}
\end{table*}

\section{Extended Related Work}
\label{sec:ext_related_work}
\paragraph{Model Explainability.}
Explainability methods aim to uncover the relevant features that influence model predictions.
The majority of works in this area emphasize local explanations of model predictions \citep{pmlr-v70-sundararajan17a, ribeiro2016should}. 
Although local explanations help to understand model behavior on specific instances, they do not provide a global understanding of model  behavior.
More recently, \citet{singh2023explaining, menon-etal-2023-mantle} proposed approaches to provide language explanations for the behavior of a model.
However, they generate explanations from a restricted set of features, either n-grams or tabular features.

\section{Additional Analysis}
\label{sec:additional_analysis}

\begin{figure}[h!]
    \centering
    \includegraphics[width=\linewidth]{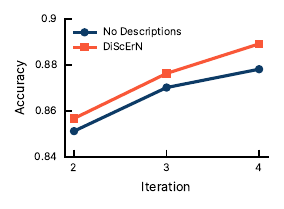}
    \vspace{-2.5em}
    \caption{
    Average \texttt{distilbert-base-uncased} accuracy when successively improved with the application of \discern and the naive augmentation (No Desc.) baseline. 
    Remarkably, the enhancement in classifier performance achieved by the No Desc. baseline in four iterations is attainable with \discern in merely three iterations.
    }
    \label{fig:increase_iterations}
\end{figure}
\paragraph{Successive use of \discern enhances the classifier even more.} 
We investigate whether the performance of a text classification model can be improved through the successive application of our framework, \discern, over multiple iterations.
We hypothesize that by iteratively refining the model using the explanations provided by \discern, and then reevaluating the model's performance, we can achieve further improvements in the classifier's accuracy. 
In Figure \ref{fig:increase_iterations}, we present the results of applying this iterative process over four successive rounds. 
The figure shows that through the repeated use of our framework and also the naive augmentation approach, the performance of the classifier continues to increase with each iteration, demonstrating the its effectiveness.
Interestingly, the enhancement in classifier performance achieved by the naive augmentation baseline in four iterations is attained with \discern in merely three iterations.

\begin{figure}
    \centering
    \includegraphics[width=\linewidth]{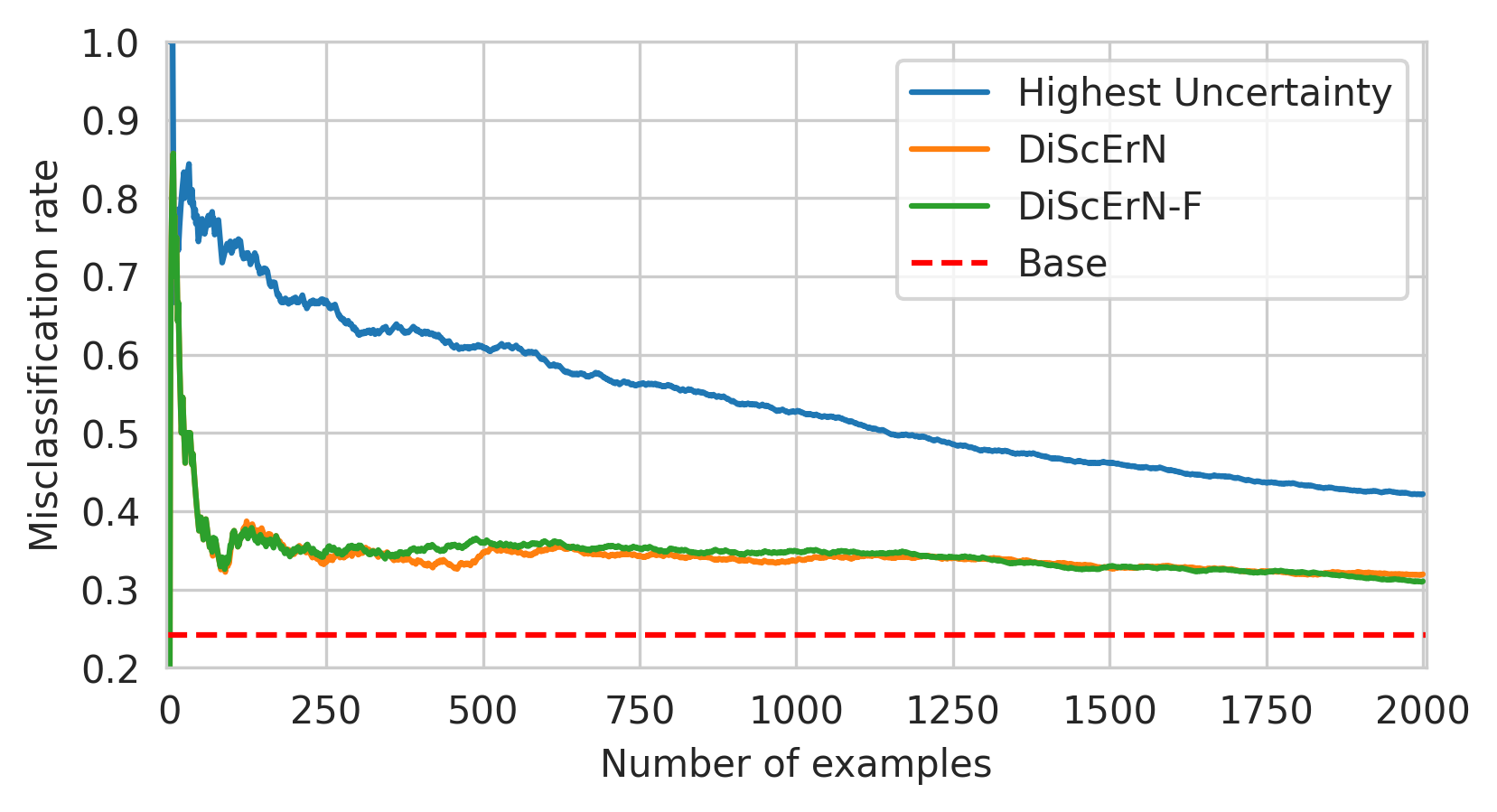}
    \caption{Misclassification rate of the top-K examples as sorted based on embedding match.}
    \label{fig:active_misclassification}
\end{figure}

\paragraph{Can \discern explanations identify the most important examples?}

In Section \ref{sec:results_and_analyses}, we present a novel active learning approach that leverages an instruction-tuned model to identify whether an example complies with a given instruction through a binary classification task.
Here, we explore an alternative retrieval-style method that utilizes large language model (LLM) embeddings to identify examples that are most likely to be misclassified. 
Specifically, we employ the LLM2Vec embeddings \citep{behnamghader2024llm2vec} to compute the semantic similarity between the instruction descriptions (queries) and the unlabeled examples (passages) in the dataset. 
By sorting the examples in decreasing order of their cosine similarity to the descriptions, we are able to plot the misclassification rate for the top-K examples as the value of K is varied.

The results presented in Figure \ref{fig:active_misclassification} demonstrate that our retrieval-based approach, using both \discern and \discernf descriptions, is capable of identifying examples that achieve a higher misclassification rate compared to randomly selecting examples. 
This finding suggests that our method is effective in identifying potentially erroneous examples. 
Interestingly, we observe that the examples selected based on the highest classifier uncertainty are consistently more challenging and exhibit a significantly higher misclassification rate.
These insights motivate future research directions that explore approaches to more consistently and effectively select harder subpopulations of data, potentially outperforming the highest classifier uncertainty approach.

\begin{table*}
    \centering
    \scalebox{0.85}{
    \begin{tabular}{l|cc|cc|cc}
    \toprule
    \textbf{Dataset $\rightarrow$} & \multicolumn{2}{c|}{\textbf{TREC (1500)}} & \multicolumn{2}{c}{\textbf{AG News (500)}} & \multicolumn{2}{c}{\textbf{Covid (4000)}} \\
    \textbf{\# Aug. Ex.} &       \textbf{500} &     \textbf{1000} &         \textbf{500} &      \textbf{1000} &        \textbf{500} &     \textbf{1000} \\
    \midrule
    Base & \multicolumn{2}{c|}{$70.85$} & \multicolumn{2}{c}{$41.2$} & \multicolumn{2}{c}{$55.1$} \\
    \midrule
    No Desc. &  $71.99_{(14.04)}$ &  $85.26_{(3.13)}$ &    $\mathbf{58.88_{(10.57)}}$ &   $61.96_{(9.47)}$ &    $50.76_{(1.42)}$ &  $46.45_{(3.07)}$ \\
    \discernf &  $72.25_{(11.95)}$ &  $86.51_{(1.98)}$ &    $58.60_{(10.35)}$ &  $64.28_{(10.67)}$ &    $\mathbf{52.35_{(1.48)}}$ &  $\mathbf{48.08_{(1.39)}}$ \\
    \textsc{\discern} &   $\mathbf{78.11_{(3.38)}}$ &  $\mathbf{88.54_{(0.90)}}$ &    $55.44_{(14.50)}$ &  $\mathbf{67.00_{(10.39)}}$ &    $51.75_{(1.76)}$ &  $47.64_{(0.91)}$ \\
    \bottomrule
    \end{tabular}
    }
    \caption{Accuracy of \texttt{roberta-large} classifier after augmenting the training set with examples that have been generated using different approaches. Numbers in brackets next to the names of the dataset indicate the number of training examples used to learn the initial classifier. \textbf{Bold} numbers indicate the best average classifier accuracy across five runs. 
    }
    \label{tab:dataset_generation_roberta_appendix}
\end{table*}

\begin{figure*}
    \centering
    \includegraphics[width=0.8\linewidth]{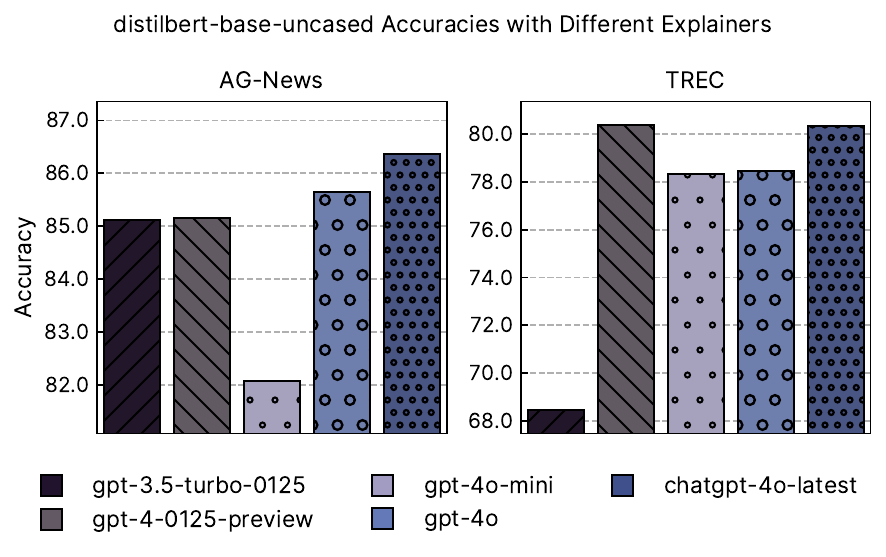}
    \caption{Accuracies post synthetic data augmentation using different language models for describing clusters. 1000 examples were augmented to the classifier based on each explanations. Results show the mean across five runs.}
    \label{fig:descriptor_ablation_extension}
\end{figure*}

\paragraph{Extending to more recent LLMs.} In Section \ref{sec:ablations} and Table \ref{tab:descriptor_ablation}, we demonstrate improvements in classifier performance facilitated by descriptions derived from the more stronger \texttt{gpt-4-turbo} model. 
Building upon this analysis, we extend our investigation to include the more recent \texttt{4o} model series, assessing their influence on the performance of the \texttt{distilbert-base-uncased} classifier across the AG-News and TREC datasets.\footnote{Note: these experiments were run after the version change mentioned in Appendix \secref{sec:training_details}.}
As can be observed in Figure \ref{fig:descriptor_ablation_extension}, using the newer variants in \texttt{gpt-4o} and \texttt{chatgpt-4o-latest}, we can obtain marked improvements over the \texttt{gpt-3.5-turbo} model. 
However, utilizing \texttt{gpt-4o-mini} results in slightly lower performance of the classifier in AG-News.
This underscores the importance of using the strongest variants of language models rather than their distilled counterparts.
Taken together, this experiments points to the potential of stronger language models contributing more accurate representations of systematic bias in text classifiers.

\section{Human Evaluation Templates}
\label{sec:human_eval_templates}
\begin{figure*}
    \centering
    \includegraphics[width=0.32\linewidth]{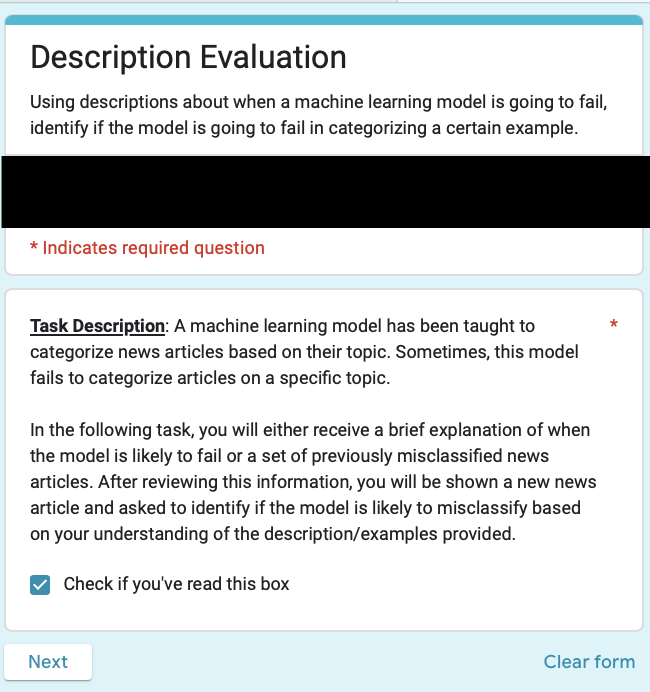}
    \hfill
    \includegraphics[width=0.32\linewidth]{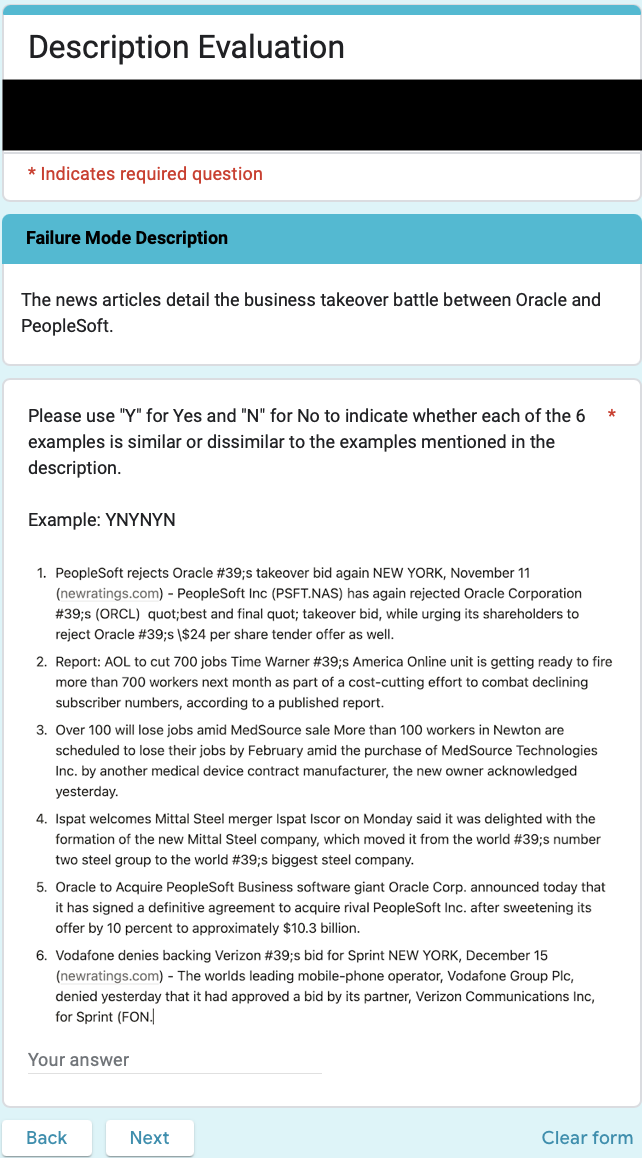}
    \hfill
    \includegraphics[width=0.32\linewidth]{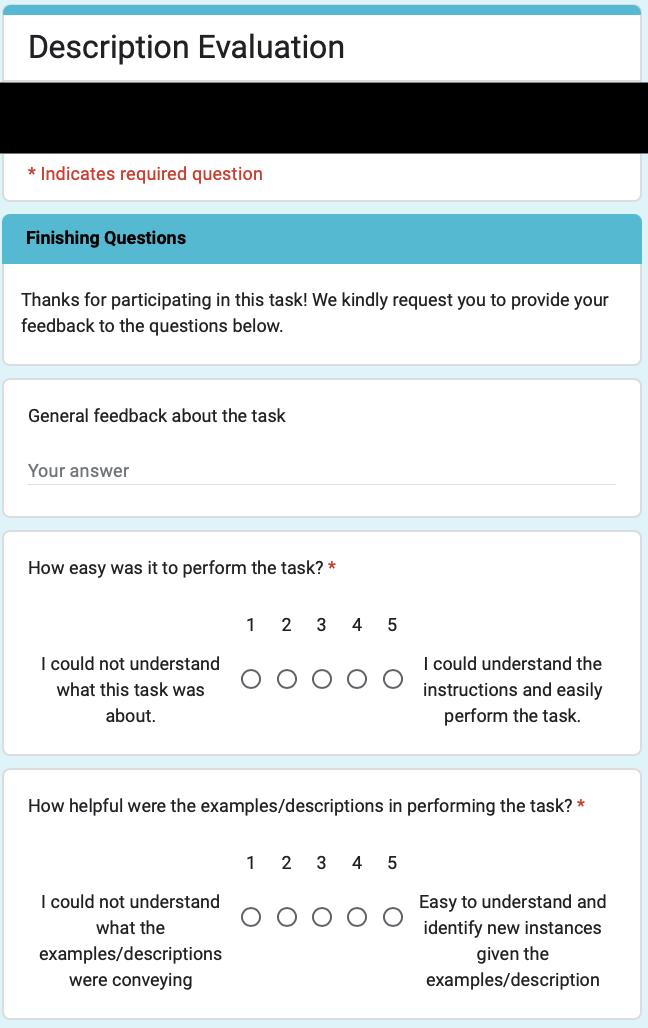}
    \caption{Example templates used for human evaluation of cluster descriptions and examples of the AGNews dataset in Section \ref{sec:results_and_analyses}.}
    \label{fig:app_human_eval}
\end{figure*}

In Figure \ref{fig:app_human_eval}, we provide screenshots of the templates used for human evaluation.

\end{document}